\documentclass[11pt]{article}
\usepackage[preprint]{acl}

\usepackage{times}
\usepackage{latexsym}
\usepackage[T1]{fontenc}
\usepackage[utf8]{inputenc}
\usepackage{microtype}
\usepackage{inconsolata}
\usepackage{graphicx}
\usepackage{booktabs}
\usepackage{amsmath}
\usepackage{amssymb}
\usepackage{multirow}
\usepackage{subcaption}
\usepackage{url}

\graphicspath{{figures/}}

\title{How Far Can You Get Without a GPU? A Systematic Benchmark of Lightweight
Hallucination Detection Across Question Answering, Dialogue, and Summarisation}

\author{Kriti Faujdar \\
  Independent Researcher \\
  \texttt{kritifaujdar@gmail.com} \\\And
  Smit Kadvani \\
  Independent Researcher \\
  \texttt{smit.kadvani@gmail.com}}

\begin{document}
\maketitle

\begin{abstract}
Hallucination detection has become a pressing requirement
for trustworthy AI deployment at scale. The most accurate detection methods depend
on GPU-intensive inference, proprietary API calls, or white-box access to the
generating model. This puts them out of reach for resource-constrained researchers
and practitioners. In this paper, we explore a practical alternative: how well can
hallucination detection perform using only lightweight, CPU-feasible methods built
on publicly available models? We systematically benchmark four such detectors,
ROUGE-L, semantic similarity, BERTScore, and a Natural Language Inference (NLI)
detector based on a FEVER-trained DeBERTa model, together with a score-level
ensemble of similarity and NLI. We evaluate them across all three tasks of the HaluEval benchmark: question
answering (QA), dialogue, and summarisation. We calibrate each method on a held-out
validation split and evaluate it on 2{,}000 test instances per task, and we report
bootstrap confidence intervals for the main results. We find that the similarity-NLI ensemble
is the most consistent method, but that absolute performance is highly
task-dependent. The ensemble ranks best on QA (F1 $=0.792$, AUC-ROC $=0.873$) and on
dialogue (F1 $=0.694$, AUC-ROC $=0.749$), where NLI is the strongest standalone
method; on summarisation every method performs near chance (AUC-ROC between
$0.469$ and $0.574$). We then ask whether that failure is intrinsic
to lightweight detection or an artifact of our single-pass design, and find it is
largely the latter. Raising the premise budget from 800 to 1600 characters lifts
summarisation AUC-ROC from $0.567$ to $0.629$, and replacing single-pass scoring
with sentence-level chunk aggregation reaches $0.683$, still on CPU and with the
same model, though at roughly twenty times the NLI inference. Summarisation remains
by far the hardest task, but our results do not support treating lightweight
detection as intrinsically unsuited to it. All experiments run on a standard laptop
CPU using public models.
\end{abstract}

\section{Introduction}
Large language models (LLMs) produce fluent text across question answering,
dialogue, summarisation, and many other natural language processing tasks
\citep{brown2020,touvron2023}. They also hallucinate. A hallucination is text that
reads as fluent and plausible but is factually incorrect or unsupported by any
available source \citep{ji2023}. In high-stakes domains such as healthcare, law,
and education, hallucinated outputs pose serious risks. This risk drives active
research into automatic hallucination detection \citep{maynez2020}.

The most accurate hallucination detection methods all depend on resources that many
researchers, educators, and practitioners do not have: GPUs, proprietary APIs, or
model internals. We discuss these methods in detail in Section~\ref{sec:lit}.

We take a constrained position. Instead of proposing another resource-intensive
detector, we ask a question. How far can hallucination detection progress using
only lightweight, CPU-feasible methods that rely on publicly available models, make
no API calls, and assume no access to the generating model? This question matters
for two reasons. First, it sets a realistic performance baseline for the large
community that works without specialised hardware. Second, it shows where
accessible methods succeed and where they break down. Papers that compare a single
expensive method against weak baselines hide this information.

We benchmark four CPU-feasible detectors that span the lexical, embedding, and
inference paradigms, ROUGE-L, semantic similarity, BERTScore, and an NLI-based
detector, together with a score-level ensemble of similarity and NLI. We evaluate them across all three tasks of
the HaluEval benchmark \citep{li2023halueval}: question answering, dialogue, and
summarisation. Our contributions are:

\begin{itemize}
\item A systematic benchmark of four CPU-feasible hallucination detectors and a
score-level ensemble across three tasks. We calibrate thresholds on held-out
validation data and evaluate on 2{,}000 test instances per task.
\item The finding that the similarity-NLI ensemble is the most consistent method,
ranking best on both QA and dialogue, while absolute performance remains strongly
task-dependent. NLI is the strongest standalone method on dialogue.
\item A diagnosis of the summarisation failure. Every method performs near chance
there, but two controlled follow-ups show this is driven largely by single-pass
truncation: enlarging the premise budget and switching to sentence-level chunk
aggregation both recover substantial performance without extra hardware, the latter
at a substantial increase in CPU inference.
\item Bootstrap confidence intervals for the main results and paired bootstrap
comparisons of AUC-ROC between the leading methods on each task, establishing which
rankings our data resolve and which it does not.
\item An error analysis of the NLI detector's failure cases, and practical
guidance for method selection under computational constraints.
\end{itemize}

\section{Related Work}
\label{sec:lit}
\citet{ji2023} survey hallucination in natural language generation. They
distinguish faithfulness (relation to a provided source) from factuality (relation
to world knowledge). HaluEval annotates candidates against a provided source, so
our study falls in the faithfulness category. Detection methods group by resource
requirement. \emph{Model-as-judge} methods prompt a strong LLM to evaluate
faithfulness. They reach high human agreement but need frontier-model access and
add per-query costs \citep{liu2023geval}. \emph{Sampling-based} methods such as
SelfCheckGPT \citep{manakul2023} measure consistency across stochastic samples.
They run in a black-box setting but need several times the generation cost.
\emph{Internal-state} methods analyse token probabilities or activations
\citep{farquhar2024}. They need white-box access, which API-only models do not
allow. \emph{Retrieval and fact-verification} pipelines decompose candidates into
atomic claims and verify each against retrieved evidence \citep{min2023factscore}.
This adds operational complexity.

The methods we evaluate belong to two lighter families. \emph{NLI-based} methods
treat the source as premise and the candidate as hypothesis. \citet{falke2019}
first repurposed NLI for summary faithfulness and found that standard NLI corpora
transferred poorly. Later work trained on the FEVER fact-verification dataset
\citep{thorne2018}. \citet{laurer2022} released DeBERTa models fine-tuned on
MultiNLI, FEVER, and Adversarial NLI, which we adopt. \citet{laban2022summac}
introduced SummaC and showed that NLI methods beat lexical metrics for
inconsistency detection. \emph{Overlap-based} metrics include ROUGE
\citep{lin2004rouge}, BERTScore \citep{zhang2020bertscore}, and Sentence-BERT
similarity \citep{reimers2019}. These are cheap to compute but can be misled when
hallucinations stay topically coherent with the source.

HaluEval \citep{li2023halueval} provides 10{,}000 samples for each of three tasks:
QA, dialogue, and summarisation. To build hallucinated candidates, the authors
prompt ChatGPT for plausible but incorrect content. Prior work has mostly reported
QA results alone. We provide a systematic comparison of lightweight methods across
all three tasks under a common calibration protocol.

\section{Dataset}
\label{sec:data}
We use all three HaluEval tasks \citep{li2023halueval} from HuggingFace
(\texttt{pminervini/HaluEval}). Each sample provides a source, a task-specific
context, and a candidate labelled faithful or hallucinated
(Table~\ref{tab:tasks}). For each task we draw 3{,}000 of the 10{,}000 samples with
seed 42, a size chosen to keep NLI inference practical on a laptop CPU while leaving
a test set large enough for stable estimates (bootstrap intervals of $\pm0.015$
to $\pm0.026$ AUC-ROC; Appendix~\ref{app:ci}). We
split them into a 1{,}000-instance validation set for threshold and ensemble-weight
calibration, and a 2{,}000-instance test set for all reported metrics. Stratified
sampling keeps the class distribution close to balanced. This split keeps every
thresholding decision off the test data.

\begin{table}[t]
\centering
\small
\caption{Structure of the three HaluEval tasks.}
\label{tab:tasks}
\begin{tabular}{@{}llll@{}}
\toprule
\textbf{Task} & \textbf{Source} & \textbf{Context} & \textbf{Candidate} \\
\midrule
QA    & Knowledge passage & Question & Answer \\
Dial. & Conversation knowl. & History & Response \\
Summ. & Document & --- & Summary \\
\bottomrule
\end{tabular}
\end{table}

\section{Methodology}
\label{sec:method}
\paragraph{Problem formulation.} Given a source $S$, optional context $C$, and
candidate $C'$, we predict $y \in \{0,1\}$, where $y=1$ denotes hallucination. Each
method produces a score $s$. A threshold $\tau$ turns the score into a binary
prediction. For every method, higher $s$ means a higher chance of hallucination;
the score definitions below make this explicit.

\paragraph{The four detectors and the ensemble.}
\textbf{ROUGE-L}: $s$ is the longest-common-subsequence F-measure between source and
candidate. \textbf{Semantic Similarity}: $s$ is the cosine similarity between
all-MiniLM-L6-v2 \citep{reimers2019} embeddings of source and candidate
($\approx$22M parameters). \textbf{BERTScore}: $s$ is the token-level contextual
F-measure with a DistilBERT backbone \citep{zhang2020bertscore}. For these three
overlap detectors we use the raw score, \emph{not} one minus it: HaluEval's
hallucinated candidates are generated to be topically plausible, so they tend to
overlap \emph{more} with the source than faithful ones do, and on QA and dialogue
higher overlap does indicate hallucination. \textbf{NLI}: the checkpoint
\texttt{MoritzLaurer/\allowbreak DeBERTa-v3-\allowbreak base-\allowbreak mnli-\allowbreak fever-\allowbreak anli}, DeBERTa-v3-base fine-tuned on
MultiNLI, FEVER, and ANLI \citep{laurer2022} ($\approx$184M parameters). The premise
is the source (truncated to 800 characters) plus context, the hypothesis is the
candidate, and $s = 1 - P(\text{entailment})$. \textbf{Ensemble}:
$s = \alpha\, s_{\text{NLI}} + (1-\alpha)\, s_{\text{sim}}$, with $\alpha$ chosen by
grid search over $\{0.1,\dots,0.9\}$ to maximise validation AUC-ROC. The selected
weights are $\alpha=0.4$ (QA), $\alpha=0.9$ (dialogue), and $\alpha=0.3$
(summarisation). These weights already show that dialogue relies almost entirely on
the NLI signal, while QA benefits from a more balanced mix. Orientation is fixed in
advance for every detector; the calibrated threshold sets only the operating point
and cannot reverse the direction of a score.

\paragraph{Input length limits.} Three of the four detectors see a truncated source.
The NLI premise is capped at 800 characters, BERTScore references at 512 characters,
and all-MiniLM-L6-v2 truncates its inputs at its default maximum of 256 tokens; only
ROUGE-L uses the full source. These limits are fixed implementation choices, not
tuned values. Section~\ref{sec:revisit} examines their effect on summarisation.

\paragraph{Calibration and metrics.} For each method, we select $\tau$ on the
validation split by maximising the Youden index ($\mathrm{TPR}-\mathrm{FPR}$) and
apply it unchanged to the test split. We report Accuracy, Precision, Recall, F1
(positive = hallucinated), and AUC-ROC, computed with scikit-learn
\citep{pedregosa2011}. Because point estimates cannot show whether a difference
between two methods is meaningful, we compute 95\% bootstrap confidence intervals for
AUC-ROC on the main results, resampling the test set with replacement over
$2{,}000$ replicates, and we compare the best and second-best method on each task by
bootstrapping their paired AUC-ROC difference on the same replicates. All experiments run on a standard laptop CPU. We
batch embeddings at 32 and NLI at 16. Our code is available at \url{https://github.com/fkriti/hallucination-detection-nli}.

\section{Results}
\label{sec:results}
\begin{figure*}[t]
\centering
\begin{subfigure}{0.49\textwidth}
  \includegraphics[width=\linewidth]{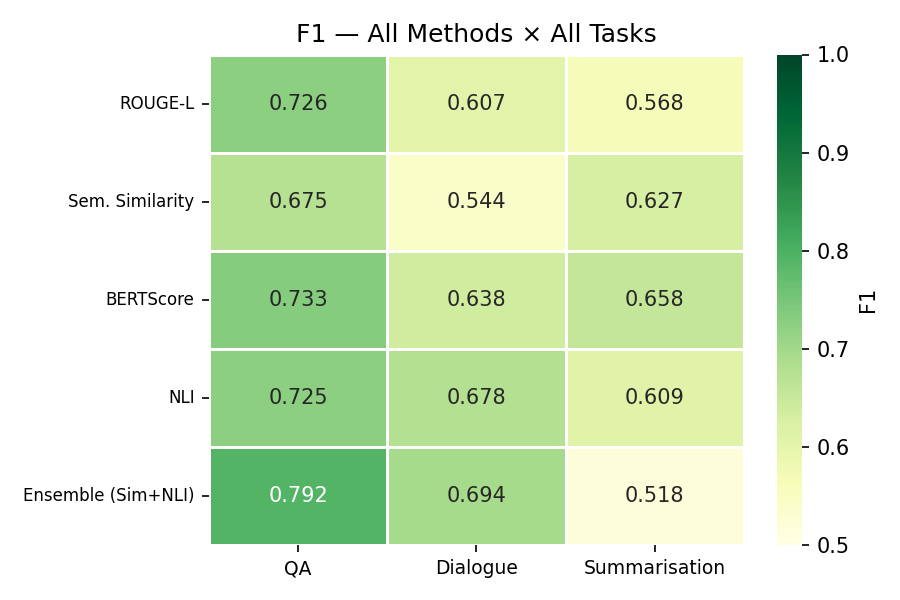}
  \caption{F1}
\end{subfigure}
\hfill
\begin{subfigure}{0.49\textwidth}
  \includegraphics[width=\linewidth]{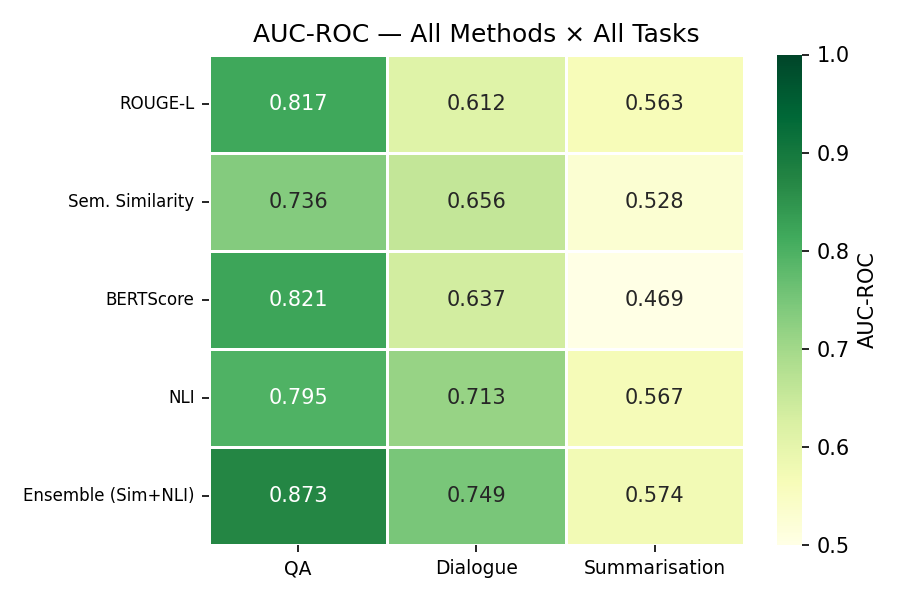}
  \caption{AUC-ROC}
\end{subfigure}
\caption{The four detectors and the ensemble across all three tasks. The task-dependent ranking and
the degradation from QA to summarisation are visible in both metrics.}
\label{fig:heatmaps}
\end{figure*}

Table~\ref{tab:main} reports all methods across all tasks. Figure~\ref{fig:heatmaps}
shows the task-dependent pattern.

\begin{table*}[t]
\centering
\renewcommand{\arraystretch}{1.15}
\setlength{\tabcolsep}{6pt}
\caption{Hallucination detection performance across the three HaluEval tasks
(test $n=2{,}000$ per task). Best result per task and metric in bold. On
summarisation, the AUC-ROC gap between the two best methods is not statistically
significant (Table~\ref{tab:pairwise}), and BERTScore's high recall and F1 there
reflect labelling nearly every candidate as hallucinated.}
\label{tab:main}
\begin{tabular}{@{}ll ccccc@{}}
\toprule
\textbf{Task} & \textbf{Method} & \textbf{Accuracy} & \textbf{Precision}
& \textbf{Recall} & \textbf{F1} & \textbf{AUC-ROC} \\
\midrule
\multirow{5}{*}{\textbf{QA}}
 & ROUGE-L             & 0.753 & 0.794 & 0.670 & 0.726 & 0.817 \\
 & Semantic Similarity & 0.655 & 0.628 & 0.730 & 0.675 & 0.736 \\
 & BERTScore           & 0.745 & 0.754 & 0.713 & 0.733 & 0.821 \\
 & NLI                 & 0.768 & \textbf{0.865} & 0.625 & 0.725 & 0.795 \\
 & Ensemble (Sim+NLI)  & \textbf{0.803} & 0.821 & \textbf{0.766}
                       & \textbf{0.792} & \textbf{0.873} \\
\midrule
\multirow{5}{*}{\textbf{Dialogue}}
 & ROUGE-L             & 0.593 & 0.577 & 0.640 & 0.607 & 0.612 \\
 & Semantic Similarity & 0.609 & 0.636 & 0.475 & 0.544 & 0.656 \\
 & BERTScore           & 0.599 & 0.572 & 0.722 & 0.638 & 0.637 \\
 & NLI                 & 0.637 & 0.600 & \textbf{0.781} & 0.679 & 0.713 \\
 & Ensemble (Sim+NLI)  & \textbf{0.672} & \textbf{0.640} & 0.756
                       & \textbf{0.694} & \textbf{0.749} \\
\midrule
\multirow{5}{*}{\textbf{Summarisation}}
 & ROUGE-L             & 0.545 & 0.531 & 0.610 & 0.568 & 0.563 \\
 & Semantic Similarity & 0.502 & 0.495 & 0.855 & 0.627 & 0.528 \\
 & BERTScore           & 0.494 & 0.492 & \textbf{0.991} & \textbf{0.658} & 0.469 \\
 & NLI                 & \textbf{0.553} & 0.533 & 0.712 & 0.609 & 0.567 \\
 & Ensemble (Sim+NLI)  & 0.552 & \textbf{0.549} & 0.490 & 0.518 & \textbf{0.574} \\
\bottomrule
\end{tabular}
\end{table*}

\paragraph{Question answering.} QA is the easiest task. The ensemble is strongest on
accuracy (.803), recall (.766), F1 (.792), and AUC-ROC (.873). It beats the best
standalone method by about 6 points on F1 and about 5 points on AUC-ROC. NLI reaches the highest
precision (.865) but low recall (.625), a conservative operating point. BERTScore
(.821 AUC) edges ROUGE-L (.817) and clearly beats semantic similarity (.736).
Contextual token overlap is therefore a stronger signal than sentence-level cosine
similarity. The ensemble's gain shows that the similarity and NLI signals
complement each other.

\paragraph{Dialogue.} Dialogue is harder. The best AUC-ROC falls to .749 (ensemble)
and .713 (NLI). The ensemble again ranks first on accuracy, precision, F1 and
AUC-ROC, while NLI is the strongest standalone method and reaches the highest
recall (.781). Entailment reasoning therefore transfers to the conversational
setting, and combining it with similarity still helps.
Lexical and embedding methods cluster at .61--.66 AUC-ROC, weaker than on QA
because responses are shorter and depend more on context.

\paragraph{Summarisation: a systematic failure mode.} Every method performs near
chance, with AUC-ROC from .469 (BERTScore) to .574 (ensemble). BERTScore's value
below .5 reflects our fixed orientation rather than a usable inverse signal:
reversing its direction gives only .532.
The high F1 for BERTScore (.658) and semantic similarity (.627) is misleading. It
comes from extreme recall (.991, .855) with near-chance precision. These methods
label almost everything hallucinated. Two factors plausibly contribute. First, a
property of the data: HaluEval summarisation hallucinations are subtle factual
edits inside long, otherwise faithful summaries of long documents, so the faithful
remainder dominates the overlap metrics and they cannot localise the inconsistent
span. Second, a property of our configuration: three of the four detectors see a
truncated source, and the NLI detector's 800-character premise in particular cannot
reach the full document. Section~\ref{sec:revisit} separates these
two explanations experimentally and shows that the second carries much of the
weight, so the figures in this row should be read as the behaviour of a
single-pass configuration rather than as a ceiling on lightweight detection.

\begin{figure*}[t]
\centering
\includegraphics[width=0.92\textwidth]{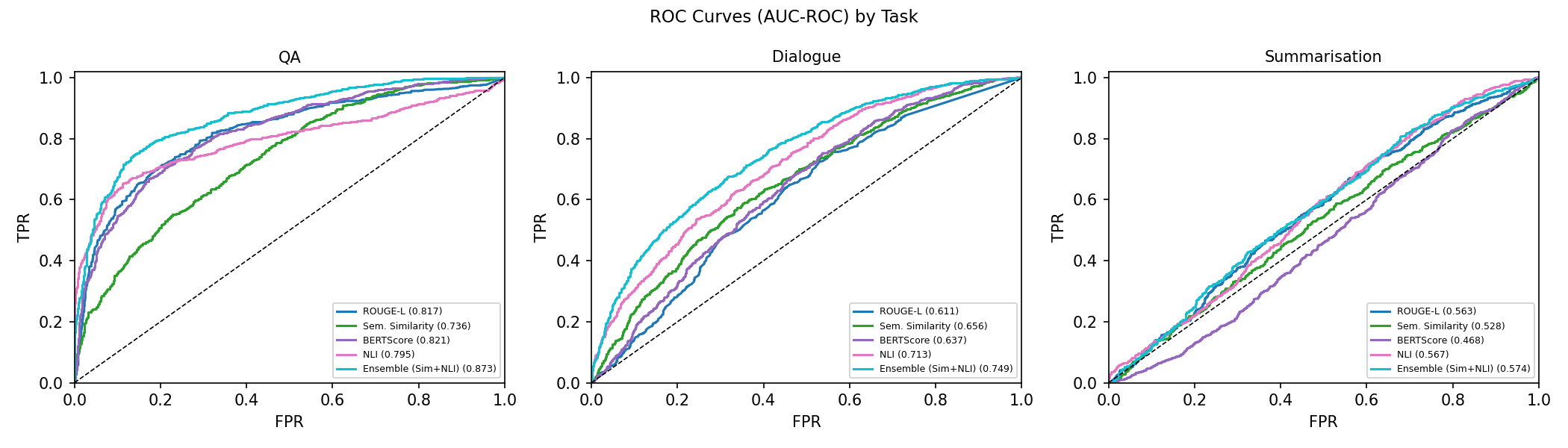}
\caption{Receiver Operating Characteristic curves for the four detectors and the ensemble on each
task. Separation from the diagonal (chance) degrades systematically from QA
(left) to summarisation (right).}
\label{fig:roc}
\end{figure*}

\paragraph{Synthesis.} On QA and dialogue the ensemble ranks first on every metric
except precision on QA and recall on dialogue, where NLI leads. These are rankings;
only the AUC-ROC gap to the runner-up is significance-tested
(Table~\ref{tab:pairwise}). Inference-based
signals, alone or combined with similarity, are therefore the most reliable choice
when lightweight detection is viable. All methods show a steep
difficulty gradient from QA (AUC up to .873) through dialogue (.749) to
summarisation (.574). The viability of GPU-free detection depends strongly on the
task. Figure~\ref{fig:roc} shows the per-task ROC curves. The gradient appears as a
progressive collapse of all curves towards the diagonal: well separated on QA,
compressed on dialogue, and indistinguishable from chance on summarisation.

\paragraph{Statistical reliability.} Point estimates alone cannot show whether one
method genuinely outperforms another, so we resample the test set with replacement
($2{,}000$ bootstrap replicates, seed 42) and recompute AUC-ROC. Full per-method
intervals appear in Appendix~\ref{app:ci}; they are narrow, typically $\pm0.02$
AUC-ROC. Table~\ref{tab:pairwise} bootstraps the paired difference
between the best method and the runner-up on each task. The ensemble's advantage is
resolved on QA and dialogue, where the interval excludes zero. On summarisation the
interval spans zero, so our data does \emph{not} establish a ranking among the
methods there. This is consistent with all of them sitting close to chance, and we
therefore avoid claiming a best method for that task. BERTScore's summarisation
interval, $[0.442, 0.493]$, lies entirely below $0.5$, but this does not make it a
useful inverse detector: orientation is fixed in advance, and reversing BERTScore's
direction gives an AUC-ROC of only $0.532$. It is near chance either way.

\begin{table}[t]
\centering
\small
\setlength{\tabcolsep}{4pt}
\caption{Paired bootstrap of the AUC-ROC gap between the best and second-best
method on each task. An interval excluding zero means the ranking is resolved.}
\label{tab:pairwise}
\begin{tabular}{@{}llcc@{}}
\toprule
\textbf{Task} & \textbf{Best vs.\ runner-up} & $\Delta$\textbf{AUC} & \textbf{95\% CI} \\
\midrule
QA     & Ensemble vs.\ BERTScore & 0.052 & $[0.032, 0.073]$ \\
Dial.  & Ensemble vs.\ NLI       & 0.036 & $[0.015, 0.057]$ \\
Summ.  & Ensemble vs.\ NLI       & 0.007 & $[-0.009, 0.022]$ \\
\bottomrule
\end{tabular}
\end{table}

\paragraph{Computational footprint.} The four detectors and the ensemble run on a
standard laptop CPU
with no GPU. Table~\ref{tab:compute} reports model size and measured throughput on
an Apple M4 laptop. The lexical and embedding methods are computationally
inexpensive. ROUGE-L
processes roughly $900$ candidates per second. The NLI detector, with its
184M-parameter model, is the bottleneck at $3.9$ candidates per second. It
scores a full 2{,}000-instance test split in about nine minutes. The ensemble adds
no model beyond NLI and similarity, so NLI dominates its cost. Even the slowest
setup stays practical for offline evaluation. The similarity-only and lexical
methods are fast enough for interactive use.

\begin{table}[t]
\centering
\small
\caption{Model size and measured CPU throughput on an Apple M4 (10 cores,
16\,GB RAM) with default PyTorch threading. Throughput is candidates scored per
second.}
\label{tab:compute}
\begin{tabular}{@{}lrr@{}}
\toprule
\textbf{Method} & \textbf{Params} & \textbf{Throughput} \\
\midrule
ROUGE-L         & ---    & $\sim$900/s \\
Sem.\ Similarity & 22M   & $\sim$58/s \\
BERTScore       & 66M    & $\sim$62/s \\
NLI             & 184M   & $\sim$3.9/s \\
Ensemble        & 206M   & $\sim$3.7/s \\
\bottomrule
\end{tabular}
\end{table}

\section{Revisiting the Summarisation Failure}
\label{sec:revisit}
The result above is easy to over-read. Reporting that every lightweight method
lands near chance on summarisation invites the conclusion that lightweight
detection is intrinsically unsuited to the task. Before drawing that conclusion we
need to rule out a simpler explanation: our own configuration. The single-pass
detector truncates the source to 800 characters, while summarisation sources have a
median length of $3{,}458$ characters (mean $3{,}905$, p90 $6{,}825$). The detector
therefore never sees roughly three quarters of a typical document. We run two
controlled follow-ups to separate the two explanations. Both keep the same NLI
model and the same CPU-only budget, so only the input handling changes.

\paragraph{Does a larger premise help?} We vary the source-truncation budget and
rerun the detector on summarisation, keeping the protocol otherwise identical
(Table~\ref{tab:trunc}). Performance is not flat. It climbs from $0.507$ AUC-ROC at
400 characters to $0.629$ at $1{,}600$, then falls back to $0.577$ at $3{,}200$. The
turning point is informative. The checkpoint is configured for at most 512
positions, and a $3{,}200$-character premise already produces a median of 629
tokens. Because DeBERTa-v3 uses relative position attention, such inputs still run
without error rather than being truncated, but they exceed the length the model is
configured for, and its entailment judgments degrade. The budget
used for our main results, 800 characters, sits well below that ceiling and leaves
roughly six AUC-ROC points unclaimed. The truncation length is thus a real
confound, though the ceiling it eventually runs into is a genuine property of the
encoder rather than a free parameter.

\begin{table}[t]
\centering
\small
\setlength{\tabcolsep}{5pt}
\caption{Summarisation performance as the NLI premise budget varies
($n=2{,}000$ test). Performance peaks below the checkpoint's configured maximum of
512 positions and degrades once the premise exceeds it.}
\label{tab:trunc}
\begin{tabular}{@{}rrccc@{}}
\toprule
\textbf{Chars} & \textbf{Med.\ tok.} & \textbf{Acc.} & \textbf{F1} & \textbf{AUC-ROC} \\
\midrule
400            & 85  & 0.520 & 0.649 & 0.507 \\
800\,(main)    & 168 & 0.553 & 0.609 & 0.567 \\
1600           & 329 & 0.598 & 0.611 & \textbf{0.629} \\
3200           & 629 & 0.530 & 0.643 & 0.577 \\
\bottomrule
\end{tabular}
\end{table}

\paragraph{Does removing truncation entirely help more?} A larger budget only
postpones the problem, because no single pass can reliably exceed that length. We
therefore test a SummaC-style alternative \citep{laban2022summac} that avoids
truncation altogether. The source is split into 800-character chunks and the
candidate into sentences; every candidate sentence is scored against every chunk;
entailment is maximised over chunks and then combined across candidate sentences,
using either the mean (the SummaC-ZS default aggregation) or the minimum, which is
more sensitive to a single unsupported sentence. Unlike SummaC-ZS, which scores
entailment minus contradiction, we use the entailment probability alone. We
deliberately reuse our own NLI model
rather than the released SummaC checkpoint, so that the comparison isolates the
aggregation strategy instead of confounding it with a different model.
Table~\ref{tab:chunkagg} reports the result on the full summarisation split against
a single-pass baseline over the same instances; that baseline reproduces the
$0.567$ of Table~\ref{tab:main} exactly. Aggregation with the mean combiner raises
AUC-ROC to $0.683$, a gain of $11.6$ points, and improves accuracy and F1 as well,
so it dominates single-pass scoring on every metric we report. It does so with no
GPU and no larger model, though it needs about twenty times as many NLI calls. The two combiners trade off as expected: the minimum is
the more sensitive detector and gives the best F1, while the mean gives the better
ranking.

\begin{table}[t]
\centering
\small
\setlength{\tabcolsep}{5pt}
\caption{Removing truncation via chunk aggregation on summarisation
($n=2{,}000$ test). All rows use the same NLI model and the same instances, so only
the input handling differs.}
\label{tab:chunkagg}
\begin{tabular}{@{}lccc@{}}
\toprule
\textbf{Method} & \textbf{Acc.} & \textbf{F1} & \textbf{AUC-ROC} \\
\midrule
Single-pass (800 chars)      & 0.553 & 0.609 & 0.567 \\
Chunk agg.\ (min combiner)   & 0.588 & \textbf{0.648} & 0.633 \\
Chunk agg.\ (mean combiner)  & \textbf{0.635} & 0.623 & \textbf{0.683} \\
\bottomrule
\end{tabular}
\end{table}

\paragraph{What this changes.} Taken together, the two experiments substantially
qualify the headline result. Summarisation is still clearly the hardest of the
three tasks, and even the best configuration we find ($0.683$) remains far below
QA ($0.873$) and dialogue ($0.749$). But the near-chance figures in
Table~\ref{tab:main} measure a naive single-pass configuration, not the capability
of lightweight methods in general. Changes that need no additional hardware
recover a large part of the gap: a larger premise budget, which lengthens each NLI
input, and chunk aggregation, which needs roughly twenty times as many NLI calls. We therefore state the finding narrowly:
single-pass, truncation-bound lightweight detection fails on long-source
summarisation, and practitioners working under CPU constraints should try
aggregation, budgeting for its extra inference, before concluding that the task is
out of reach.

\section{Error Analysis and Discussion}
\label{sec:discussion}
\paragraph{NLI errors.} On QA, false positives are short entity-only answers
(``Queen Margrethe II'', ``two''), where the brief candidate offers too little
material to confirm entailment. False negatives are fluent fabrications (``Belk was
founded in Birmingham, Alabama'') that read as locally plausible. The model is
fooled more easily by confident falsehoods than by terse truths. On dialogue, false
negatives involve plausible cross-references (e.g.\ a wrong film director) that fit
the conversational flow but contradict the knowledge. On summarisation, false
negatives differ from faithful summaries by a single altered fact that the
truncated premise cannot reach.

\paragraph{Practical guidance.} For both QA and dialogue, use the similarity-NLI
ensemble as the default: it ranks first on F1 and AUC-ROC on both tasks. Where
false positives are costly, standalone NLI gives the highest precision on QA; where
recall matters most, standalone NLI is preferable on dialogue. For summarisation,
none of the single-pass methods we test is usable; sentence-level chunk aggregation
is the best option we found, at roughly twenty times the NLI inference, and claim
decomposition is worth exploring.

\section{Conclusion}
\label{sec:conclusion}
We asked how far hallucination detection can progress without a GPU, proprietary
APIs, or access to the generating model. Across four CPU-feasible detectors, a
score-level ensemble, and three HaluEval tasks, the answer is strongly
task-dependent. A similarity-NLI ensemble is
the most consistent choice: it detects QA hallucinations well (F1 $=0.792$, AUC-ROC
$=0.873$) and also ranks first on dialogue (F1 $=0.694$, AUC-ROC $=0.749$), where
NLI is the strongest standalone method. Paired bootstrap comparisons confirm that
the ensemble's AUC-ROC lead is resolved on both tasks, while the ranking on
summarisation is not.
Every method performs near chance on summarisation (AUC-ROC $\leq 0.574$), but that
figure describes a single-pass configuration rather than a ceiling: raising the
premise budget recovers about six AUC-ROC points, and sentence-level chunk
aggregation recovers about twelve, the latter at roughly twenty times the NLI
inference and without extra hardware. Summarisation remains the hardest of the three
tasks, and the honest conclusion is narrower than the one we set out to draw.
Truncation-bound single-pass scoring fails on long-source summarisation, and our
results do not support treating lightweight detection as intrinsically unsuited to
the task. We restrict attention to
reproducible, public, CPU-only methods and report both successes and limits. This
gives practitioners without specialised hardware a realistic foundation and clear
guidance for choosing a method.

\section*{Limitations}
We study only HaluEval, whose hallucinations are synthetic and may differ from
naturally occurring ones; generalisation to hallucinations produced in the wild
remains untested. Our conclusions rest on a single sampled split with one seed, so
the bootstrap intervals we report capture test-set variability but not variation
from resampling the corpus or reshuffling the validation split, and the gains in
Section~\ref{sec:revisit} are point estimates without intervals, because we did not
retain per-instance scores for those runs. Detector orientation is fixed in advance
rather than selected on validation, which is why BERTScore can fall below chance on
summarisation. The chunk
aggregation of Section~\ref{sec:revisit} is a SummaC-style reimplementation over
our own NLI model rather than the released SummaC checkpoint, so it should be read
as evidence about aggregation strategy rather than as a benchmark of that system;
we also do not tune its chunk size, caps, or combiner, so the figures we report are
a lower bound on what aggregation can achieve. Our comparison covers lexical,
embedding, and entailment methods, but not other CPU-feasible families such as
question-generation or claim-decomposition pipelines. Finally, the ensemble is a
simple linear combination, and we leave learned stacking of all four detectors to
future work.

\section*{Acknowledgments}
This work received no external funding and was carried out on personal hardware.
The authors declare no competing interests.

\bibliography{custom}

@article{brown2020,
  title     = {Language models are few-shot learners},
  author    = {Brown, Tom B. and Mann, Benjamin and Ryder, Nick and Subbiah, Melanie and Kaplan, Jared D. and Dhariwal, Prafulla and Amodei, Dario},
  journal   = {Advances in Neural Information Processing Systems},
  volume    = {33},
  pages     = {1877--1901},
  year      = {2020}
}

@article{touvron2023,
  title   = {{LLaMA}: Open and efficient foundation language models},
  author  = {Touvron, Hugo and Lavril, Thibaut and Izacard, Gautier and Martinet, Xavier and Lachaux, Marie-Anne and Lacroix, Timoth{\'e}e and Lample, Guillaume},
  journal = {arXiv preprint arXiv:2302.13971},
  year    = {2023}
}

@article{ji2023,
  title   = {Survey of hallucination in natural language generation},
  author  = {Ji, Ziwei and Lee, Nayeon and Frieske, Rita and Yu, Tiezheng and Su, Dan and Xu, Yan and Fung, Pascale},
  journal = {ACM Computing Surveys},
  volume  = {55},
  number  = {12},
  pages   = {1--38},
  year    = {2023}
}

@inproceedings{maynez2020,
  title     = {On faithfulness and factuality in abstractive summarization},
  author    = {Maynez, Joshua and Narayan, Shashi and Bohnet, Bernd and McDonald, Ryan},
  booktitle = {Proceedings of the 58th Annual Meeting of the Association for Computational Linguistics},
  pages     = {1906--1919},
  year      = {2020}
}

@inproceedings{liu2023geval,
  title     = {{G-Eval}: {NLG} evaluation using {GPT-4} with better human alignment},
  author    = {Liu, Yang and Iter, Dan and Xu, Yichong and Wang, Shuohang and Xu, Ruochen and Zhu, Chenguang},
  booktitle = {Proceedings of the 2023 Conference on Empirical Methods in Natural Language Processing},
  pages     = {2511--2522},
  year      = {2023}
}

@inproceedings{manakul2023,
  title     = {{SelfCheckGPT}: Zero-resource black-box hallucination detection for generative large language models},
  author    = {Manakul, Potsawee and Liusie, Adian and Gales, Mark J. F.},
  booktitle = {Proceedings of the 2023 Conference on Empirical Methods in Natural Language Processing},
  pages     = {9004--9017},
  year      = {2023}
}

@article{farquhar2024,
  title   = {Detecting hallucinations in large language models using semantic entropy},
  author  = {Farquhar, Sebastian and Kossen, Jannik and Kuhn, Lorenz and Gal, Yarin},
  journal = {Nature},
  volume  = {630},
  number  = {8017},
  pages   = {625--630},
  year    = {2024}
}

@inproceedings{min2023factscore,
  title     = {{FactScore}: Fine-grained atomic evaluation of factual precision in long-form text generation},
  author    = {Min, Sewon and Krishna, Kalpesh and Lyu, Xinxi and Lewis, Mike and Yih, Wen-tau and Koh, Pang Wei and Iyyer, Mohit and Zettlemoyer, Luke and Hajishirzi, Hannaneh},
  booktitle = {Proceedings of the 2023 Conference on Empirical Methods in Natural Language Processing},
  pages     = {12076--12100},
  year      = {2023}
}

@inproceedings{li2023halueval,
  title     = {{HaluEval}: A large-scale hallucination evaluation benchmark for large language models},
  author    = {Li, Junyi and Cheng, Xiaoxue and Zhao, Wayne Xin and Nie, Jian-Yun and Wen, Ji-Rong},
  booktitle = {Proceedings of the 2023 Conference on Empirical Methods in Natural Language Processing},
  pages     = {6449--6464},
  year      = {2023}
}

@inproceedings{falke2019,
  title     = {Ranking generated summaries by correctness: An interesting but challenging application for natural language inference},
  author    = {Falke, Tobias and Ribeiro, Leonardo F. R. and Utama, Prasetya Ajie and Dagan, Ido and Gurevych, Iryna},
  booktitle = {Proceedings of the 57th Annual Meeting of the Association for Computational Linguistics},
  pages     = {2214--2220},
  year      = {2019}
}

@inproceedings{thorne2018,
  title     = {{FEVER}: A large-scale dataset for fact extraction and verification},
  author    = {Thorne, James and Vlachos, Andreas and Christodoulopoulos, Christos and Mittal, Arpit},
  booktitle = {Proceedings of the 2018 Conference of the North American Chapter of the Association for Computational Linguistics: Human Language Technologies},
  pages     = {809--819},
  year      = {2018}
}

@article{laurer2022,
  title   = {Less annotating, more classifying: Addressing the data scarcity issue of supervised machine learning with deep transfer learning and {BERT-NLI}},
  author  = {Laurer, Moritz and Van Atteveldt, Wouter and Casas, Andreu and Welbers, Kasper},
  journal = {Political Analysis},
  volume  = {32},
  number  = {1},
  pages   = {84--100},
  year    = {2024}
}

@article{laban2022summac,
  title   = {{SummaC}: Re-visiting {NLI}-based models for inconsistency detection in summarization},
  author  = {Laban, Philippe and Schnabel, Tobias and Bennett, Paul N. and Hearst, Marti A.},
  journal = {Transactions of the Association for Computational Linguistics},
  volume  = {10},
  pages   = {163--177},
  year    = {2022}
}

@inproceedings{lin2004rouge,
  title     = {{ROUGE}: A package for automatic evaluation of summaries},
  author    = {Lin, Chin-Yew},
  booktitle = {Text Summarization Branches Out},
  pages     = {74--81},
  year      = {2004}
}

@inproceedings{zhang2020bertscore,
  title     = {{BERTScore}: Evaluating text generation with {BERT}},
  author    = {Zhang, Tianyi and Kishore, Varsha and Wu, Felix and Weinberger, Kilian Q. and Artzi, Yoav},
  booktitle = {Proceedings of the 8th International Conference on Learning Representations},
  year      = {2020}
}

@inproceedings{reimers2019,
  title     = {{Sentence-BERT}: Sentence embeddings using {Siamese} {BERT}-networks},
  author    = {Reimers, Nils and Gurevych, Iryna},
  booktitle = {Proceedings of the 2019 Conference on Empirical Methods in Natural Language Processing},
  pages     = {3982--3992},
  year      = {2019}
}

@article{pedregosa2011,
  title   = {Scikit-learn: Machine learning in {Python}},
  author  = {Pedregosa, Fabian and Varoquaux, Ga{\"e}l and Gramfort, Alexandre and Michel, Vincent and Thirion, Bertrand and Grisel, Olivier and Duchesnay, {\'E}douard},
  journal = {Journal of Machine Learning Research},
  volume  = {12},
  pages   = {2825--2830},
  year    = {2011}
}

\appendix

\section{Reproducibility Details}
\label{app:repro}
All experiments use fixed random seed 42 and run on CPU. The hardware is a laptop
with an Apple M4 processor (10 cores) and 16\,GB of RAM, running macOS 26.2 and
Python 3.9.6; no GPU or accelerator is used at any stage, and we leave PyTorch
threading at its default setting. We use three public model checkpoints: \texttt{all-MiniLM-\allowbreak L6-v2} (semantic similarity),
\texttt{distilbert-\allowbreak base-\allowbreak uncased} (BERTScore backbone), and
\texttt{MoritzLaurer/\allowbreak DeBERTa-v3-\allowbreak base-\allowbreak mnli-\allowbreak fever-\allowbreak anli} (NLI). For each task we draw
3{,}000 examples. We split them into a 1{,}000-instance validation set for
threshold and ensemble-weight selection, and a 2{,}000-instance test set for
reporting. We truncate the NLI premise to 800 source characters plus up to 200
trailing context characters and BERTScore references to 512 characters, and
all-MiniLM-L6-v2 truncates its inputs at its default maximum of 256 tokens. ROUGE-L,
semantic similarity, and BERTScore use their raw scores as hallucination scores;
NLI uses $1-P(\text{entailment})$. We batch
sentence embeddings at 32 and NLI inference at 16. Thresholds maximise the
validation Youden index. The ensemble weight $\alpha$ maximises validation AUC-ROC
over $\{0.1,\dots,0.9\}$. We compute metrics with scikit-learn.

For the truncation ablation of Section~\ref{sec:revisit} we vary only the
source-character budget and leave the protocol otherwise unchanged, refitting the
threshold on validation at each setting. For chunk aggregation we split the source
into non-overlapping 800-character chunks, capped at 10 chunks, and the candidate
into sentences on terminal punctuation, capped at 6; every candidate sentence is
paired with every chunk, entailment is maximised over chunks and then combined
across candidate sentences by mean or minimum. Bootstrap intervals use $2{,}000$
replicates with seed 42.

\section{Bootstrap Confidence Intervals}
\label{app:ci}
Table~\ref{tab:ci} reports 95\% bootstrap intervals for every method and task,
computed by resampling the test set with replacement over $2{,}000$ replicates
(seed 42) and recomputing each metric on every replicate. Intervals are narrow,
which reflects the $2{,}000$-instance test splits.

\begin{table}[h]
\centering
\small
\setlength{\tabcolsep}{4pt}
\caption{95\% bootstrap confidence intervals ($2{,}000$ replicates).}
\label{tab:ci}
\begin{tabular}{@{}llcc@{}}
\toprule
\textbf{Task} & \textbf{Method} & \textbf{AUC-ROC} & \textbf{95\% CI} \\
\midrule
\multirow{5}{*}{QA}
 & ROUGE-L      & 0.818 & $[0.797, 0.838]$ \\
 & Sem.\ Sim.   & 0.736 & $[0.715, 0.757]$ \\
 & BERTScore    & 0.821 & $[0.802, 0.838]$ \\
 & NLI          & 0.795 & $[0.774, 0.816]$ \\
 & Ensemble     & 0.873 & $[0.858, 0.888]$ \\
\midrule
\multirow{5}{*}{Dial.}
 & ROUGE-L      & 0.612 & $[0.586, 0.638]$ \\
 & Sem.\ Sim.   & 0.656 & $[0.632, 0.679]$ \\
 & BERTScore    & 0.637 & $[0.614, 0.661]$ \\
 & NLI          & 0.714 & $[0.692, 0.735]$ \\
 & Ensemble     & 0.749 & $[0.729, 0.770]$ \\
\midrule
\multirow{5}{*}{Summ.}
 & ROUGE-L      & 0.563 & $[0.538, 0.588]$ \\
 & Sem.\ Sim.   & 0.528 & $[0.503, 0.553]$ \\
 & BERTScore    & 0.468 & $[0.442, 0.493]$ \\
 & NLI          & 0.567 & $[0.542, 0.592]$ \\
 & Ensemble     & 0.574 & $[0.549, 0.598]$ \\
\bottomrule
\end{tabular}
\end{table}

\section{Supplementary Figures}
\label{app:figs}

\begin{figure*}[t]
\centering
\includegraphics[width=0.92\textwidth]{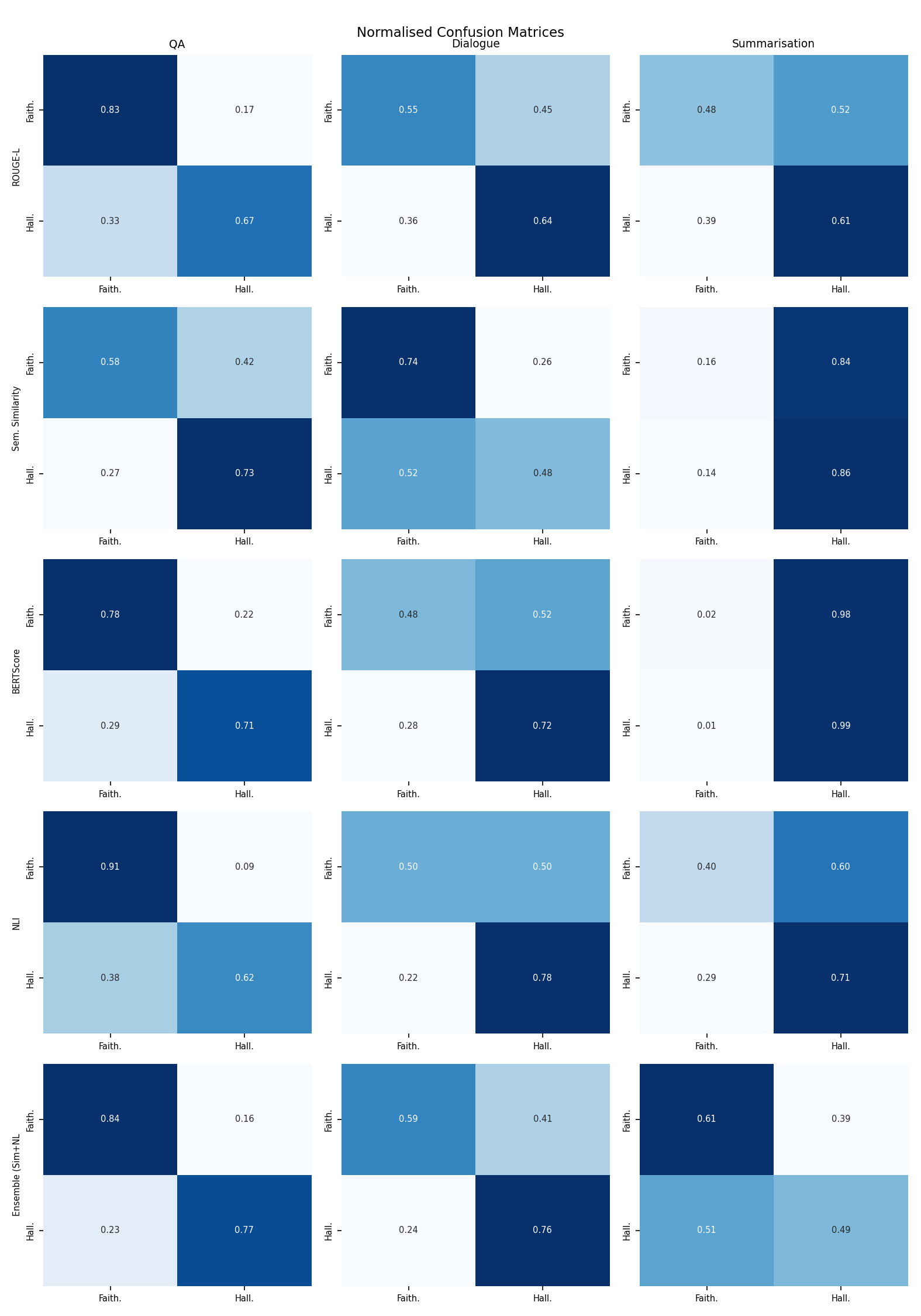}
\caption{Confusion matrices for all methods and tasks at the calibrated operating
point. The summarisation column shows the degenerate predict-almost-everything-
hallucinated behaviour of the overlap methods.}
\label{fig:cm}
\end{figure*}

\begin{figure*}[t]
\centering
\includegraphics[width=0.92\textwidth]{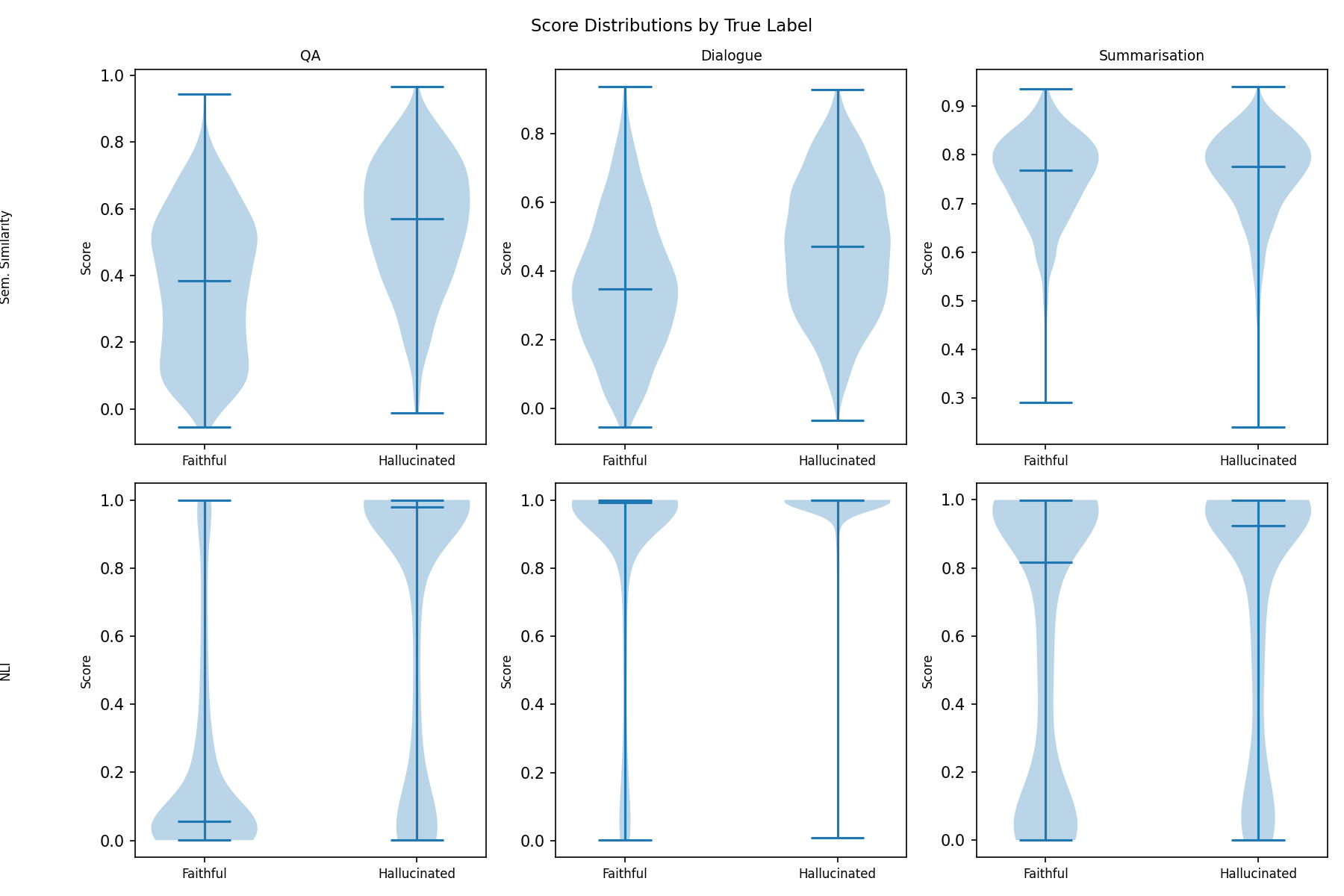}
\caption{Distributions of hallucination scores for faithful versus hallucinated
candidates. Clear separation on QA collapses to near-complete overlap on
summarisation, explaining the near-chance AUC-ROC there.}
\label{fig:dist}
\end{figure*}

\end{document}